# Construction of a Multiple-DOF Under-actuated Gripper with Force-Sensing via Deep Learning

Jihao Li[1,2], Keqi Zhu[1,2], Guodong Lu[2], I-Ming Chen[3], Huixu Dong[*1,2]

[1]Grasp Lab, Zhejiang University; [2]Robotics Institute, Zhejiang University;
[3]Robotics Research Center, Nanyang Technological University.
[*]Corresponding author：huixudong@zju.edu.cn

*Abstract*—Under-actuated robotic grippers, regarded as critical components of robotic grasping, have attracted considerable attention. However, existing under-actuated grippers emerge with several primary issues, including low payload, insufficient force sensing, small grasping force, weak grasping stability as well as high cost, hindering widespread applications. Some of these grippers can only implement a single grasping mode, thereby imposing restrictions on dimensional ranges of objects. To well relieve all relevant research gaps, we present a novel under-actuated gripper with two 3-joint fingers, which realizes force feedback control by the deep learning technique- Long Short-Term Memory (LSTM) model, without any force sensor. First, a five-linkage mechanism stacked by double four-linkages is designed as a finger to automatically achieve the transformation between parallel and enveloping grasping modes. This enables the creation of a low-cost under-actuated gripper comprising a single actuator and two 3-phalange fingers. Second, we devise theoretical models of kinematics and power transmission based on the proposed gripper, accurately obtaining fingertip positions and contact forces. Through coupling and decoupling of five-linkage mechanisms, the proposed gripper offers the expected capabilities of grasping payload/force/stability and objects with large dimension ranges. Third, to realize the force control, an LSTM model is proposed to determine the grasping mode for synthesizing force-feedback control policies that exploit contact sensing after outlining the uncertainty of currents using a statistical method. Finally, a series of experiments are implemented to measure quantitative indicators, such as the payload, grasping force, force sensing, grasping stability and the dimension ranges of objects to be grasped. Additionally, the grasping performance of the proposed gripper is verified experimentally to guarantee the high versatility and robustness of the proposed gripper. A very promising strategy combining mechanism design and artificial intelligence (AI) technology will be highly impactful on the construction of robotic grippers. *A uploaded video in YouTube*: https://youtu.be/TDyCUtxnePQ.

## I. INTRODUCTION

Grasp is a vital capability for most robots in practical applications[1, 2]. So far, various types of conventional robotic grippers have been developed for grasping objects. Yet, these robotic grippers generally include complex mechanical structures and control systems. Under-actuated robotic grippers have commonly appeared in industrial settings or logistic scenarios, being advantageous in simplifying the mechanical design, control strategy and achieving more compliant grasps [3, 4].

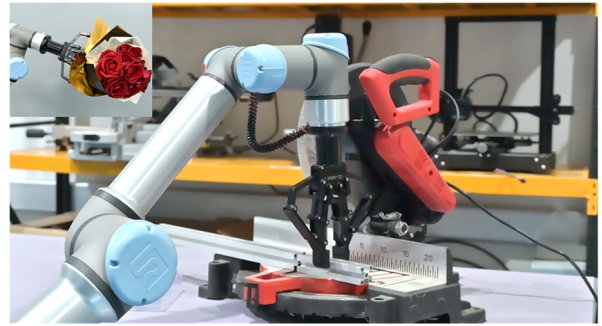

Figure. 1. The prototype of GL-Robot, which allows robust grasps in real scenarios.

Although abundant achievements have been witnessed in constructing under-actuated grippers from substantial literature available during the past few decades [3, 5], delivering prominent grasping performance and sensing contact forces remain some challenges[6], let alone developing a low-cost gripper with capabilities of high grasping payload /force /stability /force-sensing precision, large dimension ranges of objects. Generally, an under-actuated rigid gripper offers superior grasping performance compared to a soft gripper in terms of grasping payload /force /stability, especially in industrial scenarios [3, 7]. Various under-actuated rigid grippers were presented, such as the MARS and SARAH prototypes(Robotiq) [8, 9], Dong's gripper [3], and the Velo gripper[4]. Most under-actuated rigid grippers have two or three phalanges in each finger, which commonly appears in practical environments. The underlying reality is that the number of phalanges in each finger has an important effect on the grasping performance[3]. Under-actuated grippers with two phalanges in each finger have enabled rapid advancement and widespread applications such as a commercialized Robotiq85[8, 9], owing to the advantages of low-cost, simple mechanisms and control systems. Yoon *et al*.[7] proposed an underactuated gripper with a linkage mechanism to conduct a robust single pinch capability under various environmental constraints. An optimized mechanism design in the Velo gripper allows pinch and enveloping modes of compliant grasp to be transited smoothly[4]. Nevertheless, these grippers have rigorous limitations in quantitative indicators of grasping performance since it is quite difficult for them to construct form-closure using just two phalanges in each finger. To alleviate the above drawbacks, some methods typically attempt to increase phalange numbers in each finger for improving contact areas, which enables better compliant properties. For example, Dong *et al*.[3] proposed a tendon-driven compliant gripper that provides two classic grasping modes including precision and enveloping grasps. However, it only performs a single enveloping grasp mode, resulting in small ranges of objects' dimensions [3, 10, 11].

Closed-loop force controllers can for sure improve the performance of grippers. The grasping force sensing can provide the corresponding feedback for the gripper control system. Yet the force sensing methods remain an open issue and draw more attention [12]. Indirect force sensing approaches can measure contact grasping forces without any additional force sensors [13-15]. Grippers achieve mechanical force-sensing based on the mechanic's property [16], such as the snap-through phenomenon [17]. Thuruthel *et al.* [18] proposed a soft robotic gripper combining a ring and a cross-shaped structure with bistable characteristics. Generally, these force-sensing approaches construct the relationship between gripper deformations and contact forces based on virtual work [19], neural network model [13], Newton–Euler iterations [20]. However, these force-sensing methods have obvious disadvantages in terms of estimating grasping forces. Firstly, most of them can be applicable in continuum robots or soft robotic arms with long thin structures or linear-structured fin-ray grippers, which is unavailable for rigid under-actuated grippers. Secondly, the deformation-force models of these robotic arms bring many concerns about computation time and force-sensing accuracy due to their complexity and nonlinearization. Unfortunately, most existing grippers, including both commercialized and research types or rigid and soft types, are incapable of force sensing or have at most quite limited form sensing ability[13].

Motivated by the unresolved issues in the aforementioned areas, in this paper, we propose an under-actuated two-finger robotic gripper actuated by a single actuator, integrating with a force-guided precision control framework, which is named GL-Robot, as indicated in Fig.1. Each finger with three joints is in general deemed to easily construct the grasping form-closure, thereby offering a preferable grasping compliance and stability. To achieve high quantitative indicators, a five-linkage mechanism stacked by a double four-linkage mechanism is designed to conduct complex coupling and decoupling motions for amplifying the forces and torques at the fingers. In particular, when an object to be manipulated has non-contact with the proximal phalange, the gripper performs parallel grasping mode by coupling double four-linkage mechanisms; otherwise, the gripper conducts the enveloping grasp to construct form-closure owing to the mechanism descoping. The screw-nut pair serves as the primary transmission mechanism, allowing for coupling and decoupling mechanisms, which can effectively transfer the actuator power. Correspondingly, the theoretical models of kinematics and power transmission are formulated to calculate the positions and contact forces. Then, we wish to deliver a novel control architecture based on the proposed force-sensing solution by a neural network to protect fragile and deformable objects from being damaged. It is well-known that the key point for implementing a closed-loop control system is force sensing. The proposed framework of force sensing is composed of mathematical statistics (MS) outlining the uncertainty of currents and a neural network model that estimates the grasping mode to predict the phalange angle, along with the constructed mathematical models of the gripper. After obtaining this angle, the contact forces on fingers can be further calculated according to the constructed mathematical model. Results show that the proposed MS-LSTM classifier can achieve a higher recognition rate and faster inference speed; further, accurate force sensing. The gripper performance benefits from such mechanism design and force sensing strategy, as experimentally demonstrated. Compared to Robotiq85, GL-Robot shows huge advantages in quantitative indicators.

We **highlight** the **novelties** of our work. **Foremost**, our core contribution to this work is constructing an under-actuated linkage-based gripper with two 3-DOF fingers actuated by a single motor, creating potentially applicable opportunities in various fields, such as industrial settings. It is worth emphasizing that a very promising strategy combining mechanism design and artificial intelligence (AI) technology will be highly impactful on the construction of robotic grippers. The **first** novelty incorporates several new aspects. Firstly, the novelty is that we present a new mechanism stacked by four-bar linkages that conducts coupling and decoupling motions to make achievements on high grasping payload/ force/stability and large dimension ranges of objects. Secondly, a mathematical model of the proposed mechanism is established to calculate the position and contact force of phalanges. In terms of the **second** novelty, the work provides the first solution by deep learning model integrated with mathematical statistics and the constructed mechanism model to realize force sensing without any force sensor, to the best of our knowledge, which can be easily generalized to force control of other compliant grippers. We experimentally demonstrate that GL-Robot presents an excellent grasping performance on quantitative indicators in physical environments, which is attributed to the **third** contribution. It is possible for GL-Robot to be commercialized like Robotiq85, bringing out the **fourth** contribution.

## II. MODELING AND ANALYSIS

The schematic of the proposed under-actuated finger is illustrated in Fig.2. This finger consists of three consecutive phalanges and two driving transmission linkages. The actuation torque $\tau_a$ at the joint $O_a$ brings to the associated rotation for driving three phalanges via the multiple-linkage mechanism. The rotation joint $O_3$ is equipped with one torsional spring, which remains the default state without external forces or torques. In particular, the angle between the intermediate and distal phalanges has a default intersection angle $180°$. Due to the mechanical limits, the ranges of rotation angle $\theta_2$ are located at $[0°, 90°]$ while the proximal joint can realize a rotation range between $20°$ and $110°$ (see Fig.3).

### A. Mechanism Design of GL-Robot

The mechanism design of GL-Robot is introduced in detail. To enable an under-actuated gripper to perform parallel and enveloping grasps, we aspect that the under-actuated grasp consists of two phases for parallel and force grasps. First, before no external contact occurs at the proximal phalange, the intermediate and proximal phalanges behave as the entire body since the torsional spring installed on the joint between these two phalanges limits a relative free rotation. The preloading of the torsional spring can prevent any undesired motions that are brought by gravity and inertia during the grasping period. Second, if the object to be grasped is small, the gripper will conduct a parallel grasp (see Fig.2-A). The distal phalange touches it while this phalange still keeps the same state due to the mechanical limitations. If the finger grips a big object, the finger tends to conduct a force grasp by enveloping the object, as depicted in Fig.2(B). The proximal phalange stops the rotation once it makes physical contact with this target. Then, the torque from the motor overcomes the preloading of the torsional springs located on the joints so that the intermediate and distal phalanges continue rotating relative to the proximal phalanx against the torsional springs until these two phalanges consecutively have contact with the object. The stiffness of torsional springs installed on the joints should be designed as small as possible, but sufficiently big to prevent undesired motions during the free open-close period. Moreover, it is noted that when the gripper releases the object, stored loads in these torsional springs allow the phalanges to return to their initial configuration. Such sequence movements are generated by one actuator.

As mentioned above, the designed gripper needs to realize parallel grasp for small objects and force grasp via enveloping grasp to cover objects with relatively big sizes in real scenarios. It is easier to achieve grasping stability for a gripper with multiple phalanges than one with two phalanges [3]. Therefore, GL-Robot includes two 3-DOF under-actuated fingers, whereas there is just one actuator. The torsional spring and mechanical stoppers are employed to configure relative positions of linkages consisting of fingers at the initial state. The prototype and the schematic diagrams of GL-Robot are shown in Fig.3. The gripper with two 3-phalange fingers is driven by a single brushless actuator of "EYOU-proServo" with a nominal torque of 1.2Nm. A lead-screw and screw-nut transmission with a wire pitch of 1 mm enables the maximum output torque to be increased to 41.76 Nm. Its non-back-drivability allows the phalanges to maintain the contact positions of an object, although the actuator is powered off. Two torsion springs commercially available are installed in two joints of distal and middle phalanges to ensure the initial states of fingers owing to the stored energy of the springs. For each finger, it has a total stiffness coefficient of 0.038 Nm/deg since the stiffness coefficient of a torsion spring is 0.019 Nm/deg. The electronic driver for the motion control is mounted at the bottom of the palm, as shown in Fig.3(A). A parallelogram-linkage mechanism has been widely applied to obtain a constant orientation of the proximal phalanx that once contacts an object. The finger is constructed with the serial bars and the parallelogram linkages sharing middle phalanges as common links. A transmission link is directly connected to the distal phalange. When the distal phalange encounters the physical constraint while other phalanges have no contacts, all the points on the distal phalange have the same linear velocity vector.

The movement of the distal phalange needs to be decoupled from another two phalanges on the same finger to reduce control complexity and increase grasping dexterity, as shown in Fig.3(B). The design is that we stack one five-bar mechanism ($ABCDEF$) over one fixed parallelogram ($DEIJ$) and the other flexible parallelogram ($EFGH$) to construct the finger mechanism. That is, double parallelograms are in the six-bar mechanism. Without external contacts, the proximal phalange ($PDC$) and intermediate phalange ($DEIJ$) are passively coupled with each other by the torsional springs and mechanical limitations. For a parallel grasping mode, the moment generated by the torsional springs at the joint $O_3$ pushes the linkage ($DC$) onto the linkage ($DJ$). Furthermore, when a reaction force exerts on the distal phalange brings out a moment, this moment enables the linkage ($DC$) to press on the linkage ($DJ$) such that the angle of the linkage ($PD$) maintains vertically along the horizontal direction. Moreover, owing to the fixed parallelogram inside the five-bar mechanism, the partial grasping reaction force can be supported by the linkage ($IJ$), which reduces the actuation torque. In terms of force grasp, when a reaction force occurs on the proximal phalange, the generated moment allows the linkage ($EI$) to be separated from the linkage ($EH$). Similarly, the fixed parallelogram ($DEIJ$) as the intermediate phalange is also separated from the distal phalange when a reaction force on the intermediate phalange brings out a moment. Thus, the angle of the linkage ($PD$) is changing such that the finger adapts to the object's shape.

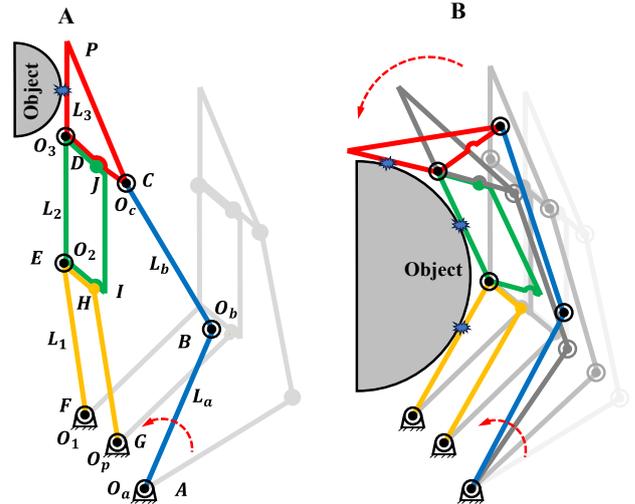

Figure. 2. A novel mechanism stacked by a double four-linkage mechanisms. The parallel grasp (A) and the enveloping grasp (B). The gray schematic diagram illustrates the state of the finger during the period of the finger performing grasps at previously sampled time. $O_i (i = a, b, c, 1, 2, 3)$ represents the number of rotation joints; $L_i, (i = 1,2,3)$ denote the rotation angle of the length of the the $i$-th phalange, respectively; $L_i (i = a, b)$ indicates the driving transmission linkage. $\tau_a$ is the actuator torque.

### B. Inverse Kinematics of GL-Robot

Here we construct the inverse kinematics model through the position of the fingertip $P$ to find the actuation angle $\theta_a$ for the parallel and force grasps. As illustrated in Fig.3(B), the vector-loop equation is provided as

$$\overrightarrow{FE} + \overrightarrow{ED} + \overrightarrow{DC} = \overrightarrow{FA} + \overrightarrow{AB} + \overrightarrow{BC} \quad (1)$$

In terms of parallel grasp, the angle $\alpha$ between the linkage ($PD$) and the $x$-axis is a constant and the angle $\beta (\beta = \angle PDC)$ is fixed. $\theta_a$ can be readily calculated by Eq.(1). As for the force grasp, the point $P(p_x, p_y)$ is provided as

$$\begin{aligned} p_x &= L_1 \cos \theta_1 + L_2 \cos \beta + L_3 \cos \alpha \\ p_y &= L_1 \sin \theta_1 + L_2 \sin \beta + L_3 \sin \alpha \end{aligned} \quad (2)$$

To determine $p_y$, we eliminate $\theta_1$ by

$$(p_x - L_2 \cos \beta - L_3 \cos \alpha)^2 + (p_y - L_2 \sin \beta - L_3 \sin \alpha)^2 = L_1^2 \quad (3)$$

Solve Eq.(3) for $p_y$ to yield

$$\begin{aligned} p_y^2 + p_x^2 - \kappa_1 p_x - \kappa_2 p_y + \kappa_3 &= 0 \\ \kappa_1 &= 2(L_2 \cos \beta + L_3 \cos \alpha) \\ \kappa_2 &= 2(L_2 \sin \beta + L_3 \sin \alpha) \\ \kappa_3 &= L_2^2 + L_3^2 - L_1^2 + 2L_2 L_3 \cos(\alpha - \beta) \end{aligned} \quad (4)$$

Thus,

$$p_y = \frac{1}{2} \left( \kappa_2 \pm \sqrt{\kappa_2^2 - 4(p_x^2 - \kappa_1 p_x + \kappa_3)} \right) \quad (5)$$

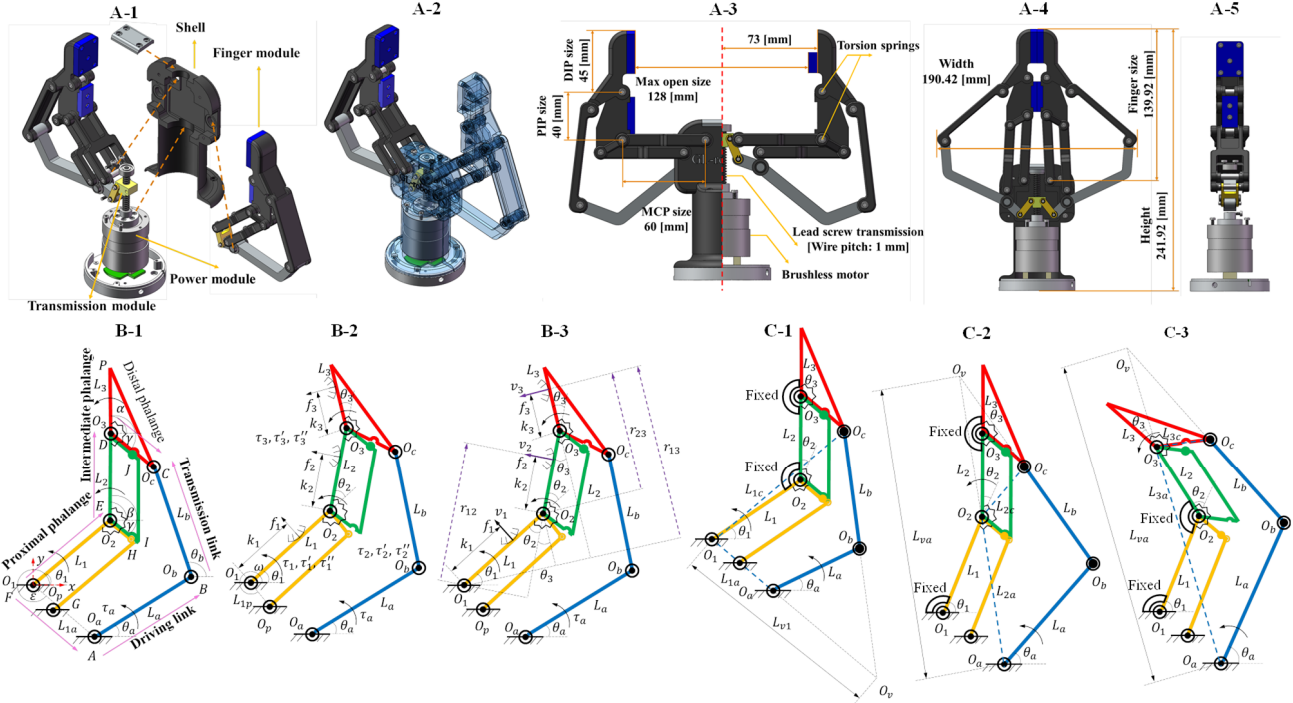

Figure. 3. The prototype of GL-Robot (A), the schematic diagrams of the gripper's parameters, kinematics and force transmission(B) as well as phalange's velocities (C). $\theta_a$ is the actuation angle; $\theta_i, k_i, f_i, \tau_i, \tau_i', \tau_i'' (i = 1,2,3)$ denote the rotation angle of the $i$-th joint, the segment between the $i$-th joint and contact point, the contact force, the input torque, the output torque and the contact torque of the the $i$-th phalange, respectively; $O_v$ is the instantaneous center of velocity for four-bar linkage $O_a O_b O_c O_i$ is the intersection point of two lines $O_b O_c$ and $O_a O_i$; $\alpha, \theta_a, \theta_b, \theta_1$ are measured from the x-axis; $\varepsilon$ indicates the angle between the x-axis and $FA$, respectively; $\gamma$ depicts the angle between $DP$ and $DC$.

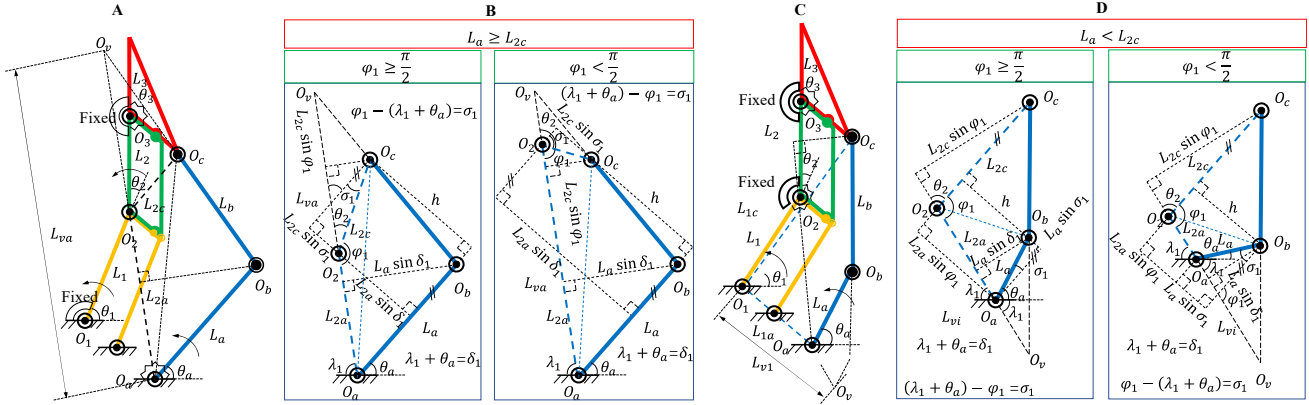

Figure. 4. The schematic diagrams of two classifications of the instantaneous center $O_v$ of velocity for GL-Robot (A, C). Calculating four cases of the instantaneous center $O_v$ of velocity for four-bar linkage $O_a O_b O_c O_i$ (B, D).

The solution with the positive sign can be chosen owing to the mechanical constraint (the other solution exists inside the six-bar mechanism). Similarly, the point $C(c_x, c_y)$ is determined by

$$c_x = p_x - L_3 \cos \alpha + L_{3C} \cos(\alpha - \gamma)$$
$$c_y = p_y - L_3 \sin \alpha + L_{3C} \sin(\alpha - \gamma) \quad (6)$$

Moreover, the position of the point $C$ can be calculated by the following vector-loop equation as

$$\vec{FC} = \vec{OA} + \vec{AB} + \vec{BC} \quad (7)$$

This can be expressed by

$$c_x = L_{1a} \cos \varepsilon + L_a \cos \theta_a + L_b \cos \theta_b$$
$$c_y = L_{1a} \sin \varepsilon + L_a \sin \theta_a + L_b \sin \theta_b \quad (8)$$

Eliminating $\theta_b$, we have

$$(c_x - L_{1a} \cos \varepsilon - L_a \cos \theta_a)^2 + (c_y - L_{1a} \sin \varepsilon - L_a \sin \theta_a)^2 = L_b^2 \quad (9)$$

Furthermore, the above equation can be expressed as

$$\rho_1 \cos \theta_a + \rho_2 \sin \theta_a + \rho_3 = 0$$
$$\rho_1 = 2L_{1a}(L_{1a} \cos \varepsilon - c_x)$$
$$\rho_2 = 2L_a(L_{1a} \sin \varepsilon - c_y)$$
$$\rho_2 = (c_x - L_{1a} \cos \varepsilon)^2 + (c_y - L_{1a} \sin \varepsilon)^2 + L_a^2 - L_b^2 \quad (10)$$

Setting $t = \tan \frac{\theta_a}{2}$, we rearrange the above equation as

$$(\rho_3 - \rho_1)t^2 + 2\rho_2 t + \rho_1 + \rho_3 = 0 \quad (11)$$

Therefore, the actuation angle $\theta_a$ can be calculated by

$$\theta_a = 2\tan^{-1}\left(\frac{-\rho_2 \pm \sqrt{\rho_2^2 - (\rho_3 - \rho_1)(\rho_1 + \rho_3)}}{(\rho_3 - \rho_1)}\right) \quad (12)$$

The solution with the positive sign is used for determining the grasp space. When $\rho_2^2 < \rho_3^2 - \rho_1^2$, no solution for $\theta_a$ exists. The inverse kinematics model can be used in the position-feedback control of GL-Robot.

### C. Contact Forces on Phalanges

To analyze the grasping configuration for the finger, the model that indicates the actuation torque and the contact force, torques on the finger must be achieved [3]. The two matrices are applied to establish a mathematical relationship, including the Jacobian matrix that maps the actuation torque to the contact forces and the transmission matrix that describes the geometrical transmission theories of linkages. Here we ignore the friction between a finger and an object, considering the worst grasping scenario where there is no friction affecting the grasp stability. If realizing a stable grasp without friction, GL-Robot can offer a better grasping performance when the friction rises. Therefore, the contact forces at contact points are normal in the Jacobian matrix.

There is only one actuator driving all the linkages to adapt to the target object. To obtain the contact forces, we consider the static-equilibrium model via equating the input and output virtual powers, which is provided as

$$t^T \omega_a = F^T v = \tau^T \dot{\theta} \quad (13)$$

where $t$ represents the input torque vector from the actuator and torsional springs installed on the rotation joints, as shown in Fig.3(B); $\omega_a$ denotes the angular velocity vector of the actuator and torsional springs; $F$ is the contact force and the positive directions of these contact forces point to the object; $v$ indicates the projected velocity vector of the contact points; $\tau$ and $\dot{\theta}$ are the vectors of the torques and angular velocities of phalange joints, respectively. The above vectors are given as,

$$t = \begin{bmatrix} \tau_a \\ \tau_1 = -K_1 \Delta\theta_1 \\ \tau_2 = -K_2 \Delta\theta_2 \\ \tau_3 = -K_3 \Delta\theta_3 \\ \ldots \\ \tau_n = -K_n \Delta\theta_n \end{bmatrix}, \omega_a = \begin{bmatrix} \dot{\theta}_a \\ \dot{\theta}_1 \\ \dot{\theta}_2 \\ \dot{\theta}_3 \\ \ldots \\ \dot{\theta}_n \end{bmatrix}, F = \begin{bmatrix} f_1 \\ f_2 \\ f_3 \\ \ldots \\ f_n \end{bmatrix}, v = \begin{bmatrix} v_1 \\ v_2 \\ v_3 \\ \ldots \\ v_n \end{bmatrix},$$

$$\tau = \begin{bmatrix} \tau'_1 \\ \tau'_2 \\ \tau'_3 \\ \ldots \\ \tau'_n \end{bmatrix}, \dot{\theta} = \begin{bmatrix} \dot{\theta}_1 \\ \dot{\theta}_2 \\ \dot{\theta}_3 \\ \ldots \\ \dot{\theta}_n \end{bmatrix} \quad (14)$$

in which $\tau_a$ represents the actuation torque exerted at the base joint $O_a$ from one motor; $\tau_i$ is the torque from the torsional spring located in the joint $O_i (1 \leq i \leq n)$; $n(n \geq 1)$ is the number of joints on the phalanges; $\Delta\theta_i$ denotes the difference between the current angle and the initial one of the joint; $K_a$ and $K_n$ are the stiffness coefficients of the torsional springs installed in the actuation joint and phalange joint, respectively. Note that $K_a$ or $K_i$ becomes zero when the rotation joint is without the torsional joint; $\dot{\theta}_a$ and $\dot{\theta}_i$ indicate the angular velocities at the actuation joint and $i^{\text{th}}$ phalange joint, respectively; $v_i$ denotes the velocity of the $i$-th contact point; $\tau'_i$ and $f_i$ are respectively the output torque and the corresponding contact force of the $i$-th rotational joint of the phalange. Thus, these vectors can be defined as

$$t = \begin{bmatrix} \tau_a \\ \tau_2 \\ \tau_3 \end{bmatrix}, \omega_a = \begin{bmatrix} \dot{\theta}_a \\ \dot{\theta}_2 \\ \dot{\theta}_3 \end{bmatrix}, F = \begin{bmatrix} f_1 \\ f_2 \\ f_3 \end{bmatrix}, v = \begin{bmatrix} v_1 \\ v_2 \\ v_3 \end{bmatrix},$$

$$\tau = \begin{bmatrix} \tau'_1 \\ \tau'_2 \\ \tau'_3 \end{bmatrix}, \dot{\theta} = \begin{bmatrix} \dot{\theta}_1 \\ \dot{\theta}_2 \\ \dot{\theta}_3 \end{bmatrix} \quad (15)$$

where $\tau_2$ and $\tau_3$ represent the passive torques of the torsional springs located at the $o_2$ and $o_3$ joints, respectively; thus, $\tau_2 = -K_2 \Delta\theta_2$ and $\tau_3 = -K_3 \Delta\theta_3$.

Furthermore, the projected velocities $v = [v_1, v_2, v_3, \cdots, v_n]^T$ can be obtained by a product of a Jacobian matrix $J$ and the derivative vector of the phalanx joint coordinates $\dot{\theta} = [\dot{\theta}_1, \dot{\theta}_2, \dot{\theta}_3, \cdots, \dot{\theta}_n]^T$, namely,

$$v = J\dot{\theta} \quad (16)$$

Similarly, the transmission matrix $T$ that describes the transmission mechanism applied in a finger can map the angular velocities $\dot{\theta}$ of joints on the phalanges to the actuation angular velocities $\omega_a$ from the actuator and torsional springs, i.e.:

$$\omega_a = T\dot{\theta} \quad (17)$$

Thus, combining Eqs.(12,16,17), we can obtain the following equation,

$$\begin{aligned} \tau &= T^T t \\ F &= J^{-T} \tau \\ F &= J^{-T} T^T t \end{aligned} \quad (18)$$

To calculate the contact forces, we use the geometrical analytical method to express the Jacobian matrix in a lower triangular form as follows,

$$J = \begin{bmatrix} k_1 & 0 & 0 & \cdots & 0 \\ r_{12} & k_2 & 0 & \cdots & 0 \\ r_{13} & r_{23} & k_3 & \cdots & 0 \\ \ldots & \ldots & \ldots & \ldots & \ldots \\ r_{1n} & r_{2n} & r_{3n} & \cdots & k_n \end{bmatrix}$$

with $r_{ii} = k_i$. As illustrated in Fig.3(B), if $n = 3$, we can have the following equation,

$$\begin{aligned} r_{12} &= k_2 + L_1 \cos\theta_2 \\ r_{13} &= k_3 + L_1 \cos(\theta_2 + \theta_3) + L_2 \cos\theta_3 \\ r_{23} &= k_3 + L_2 \cos\theta_3 \end{aligned} \quad (19)$$

Therefore, from Eq.(16), the velocities of contact points on the phalanges can be provided as

$$\begin{aligned} v_1 &= k_1 \dot{\theta}_1 \\ v_2 &= (k_2 + L_1 \cos\theta_2)\dot{\theta}_1 + k_2 \dot{\theta}_2 \\ v_3 &= [k_3 + L_1 \cos(\theta_2 + \theta_3) + L_2 \cos\theta_3]\dot{\theta}_1 + \\ &\quad (k_3 + L_2 \cos\theta_3)\dot{\theta}_2 + k_3 \dot{\theta}_3 \end{aligned} \quad (20)$$

The above Jacobian matrix can be used in both compliant mechanism and fully actuated mechanism as this matrix is related to dimension parameters rather than transmission system. Conversely, the transmission matrices of compliant and fully actuated fingers are different since they have different transmission systems. In terms of the transmission matrix $T$, it is observed that $T$ is the identity matrix for fully actuated fingers. As shown in Fig.3(C), for a compliant finger, the geometrical analytic approach is applied to calculate the transmission matrix $T$ with $(n + 1) \times n$ elements. The expanded form of Eq.(17) is given as

$$\begin{bmatrix} \dot{\theta}_a \\ \dot{\theta}_1 \\ \dot{\theta}_2 \\ \dot{\theta}_3 \\ \ldots \\ \dot{\theta}_n \end{bmatrix} = \begin{bmatrix} X_1 & X_2 & X_3 & \cdots & X_n \\ 1 & 0 & 0 & \cdots & 0 \\ 0 & 1 & 0 & \cdots & 0 \\ 0 & 0 & 1 & \cdots & 0 \\ \ldots & \ldots & \ldots & \ldots & \ldots \\ 0 & 0 & 0 & \cdots & 1 \end{bmatrix} \begin{bmatrix} \dot{\theta}_1 \\ \dot{\theta}_2 \\ \dot{\theta}_3 \\ \ldots \\ \dot{\theta}_n \end{bmatrix} \quad (21)$$

when $n = 3$, referring to Fig.3(C), we have the following equation

$$\begin{bmatrix} \dot{\theta}_a \\ \dot{\theta}_2 \\ \dot{\theta}_3 \end{bmatrix} = \begin{bmatrix} X_1 & X_2 & X_3 \\ 0 & 1 & 0 \\ 0 & 0 & 1 \end{bmatrix} \begin{bmatrix} \dot{\theta}_1 \\ \dot{\theta}_2 \\ \dot{\theta}_3 \end{bmatrix} \quad (22)$$

Thus,

$$\dot{\theta}_a = X_1 \dot{\theta}_1 + X_2 \dot{\theta}_2 + X_3 \dot{\theta}_3 \quad (23)$$

To determine $X_i$ mapping $\dot{\theta}_a$ to $\dot{\theta}_i$ based on the principle of virtual work, we need to lock other phalanges for reversing a rotational degree of freedom (DOF). The two locked phalanges behave as the entire linkage, as shown in Fig.3(C). $X_i$ is calculated by

$$X_i = \frac{\dot{\theta}_a}{\dot{\theta}_i}, i = 1,2,3 \quad (24)$$

Kennedy's Theorem [21] is used for calculating the angular velocity $\dot{\theta}_i$ of the $i$-th phalange, as illustrated in Fig.3(C). The instantaneous center $O_v$ of velocity for four-bar linkage $O_a O_b O_c O_i$ is the intersection point of two lines $O_b O_c$ and $O_a O_i$. Specifically, if $L_a \geq L_{ic}$, $O_v$ is on the extended line of the vector $\overrightarrow{O_a O_i}$; otherwise, the extended line of the vector $\overrightarrow{O_i O_a}$ goes through $O_v$. Thus, the following equation is provided as

$$\frac{\dot{\theta}_a}{\dot{\theta}_i} = \frac{L_{va}-L_{ia}}{L_{va}}, L_a \geq L_{ic}$$
$$\frac{\dot{\theta}_a}{\dot{\theta}_i} = \frac{L_{vi}}{L_{vi}-L_{ia}}, L_a < L_{ic} \quad (25)$$

where $L_{ia}(i = 1,2,3)$ is the length between the actuation joint and the $i$-th joint of phalange; $L_{va}$ or $L_{vi}$ is a side length of the triangle formed by two lines $O_b O_c$ and $O_a O_i$. To calculate $L_{va}$ under the condition of $L_a \geq L_{ic}$ and $\varphi_i \geq \frac{\pi}{2}$ (see Fig.4), we employ the geometrical area method as follows,

$$S_{\triangle O_a O_b O_c} = S_{\triangle O_v O_a O_b} - S_{\triangle O_v O_a O_c}$$
$$S_{\triangle O_a O_b O_c} = \frac{1}{2} L_{va} L_a \sin(\lambda_i + \theta_a) - \frac{1}{2} L_{va} L_{ic} \sin \varphi_i \quad (26)$$
$$S_{\triangle O_a O_b O_c} = \frac{1}{2} L_a h$$
$$S_{\triangle O_a O_b O_c} = \frac{1}{2} L_a [L_{ia} \sin(\lambda_i + \theta_a) + L_{ic} \sin(\varphi_i - (\lambda_i + \theta_a))] \quad (27)$$

Combining Eqs.(26-27), we have $L_{va}$ as

$$L_{va} = \frac{L_a[L_{ia}\sin(\lambda_i+\theta_a)+L_{ic}\sin(\varphi_i-(\lambda_i+\theta_a))]}{L_a\sin(\lambda_i+\theta_a)-L_{ic}\sin\varphi_i} \quad (28)$$

For the two cases where $\varphi_i \geq \frac{\pi}{2}$ and $\varphi_i < \frac{\pi}{2}$, they have the same $L_{va}$, as shown in Fig.4. Satisfying the condition of $L_a \geq L_{ic}$ and $\varphi_i \geq \frac{\pi}{2}$, the same geometrical area method is used for determining $X_i$ as

$$S_{\triangle O_b O_c O_i} = S_{\triangle O_v O_c O_i} - S_{\triangle O_v O_b O_i}$$
$$S_{\triangle O_b O_c O_i} = \frac{1}{2} L_{vi} L_{ic} \sin \varphi_i - \frac{1}{2} L_{vi} L_a \sin(\lambda_i + \theta_a) \quad (29)$$
$$S_{\triangle O_b O_c O_i} = \frac{1}{2} L_{ic} h$$
$$S_{\triangle O_b O_c O_i} = \frac{1}{2} L_{ic}[L_{ia}\sin\varphi_i - L_a \sin((\lambda_i + \theta_a) - \varphi_i)] \quad (30)$$

Therefore, we can obtain $L_{vi}$ as

$$L_{vi} = \frac{L_{ic}[L_{ia}\sin\varphi_i - L_a\sin((\lambda_i+\theta_a)-\varphi_i)]}{L_{ic}\sin\varphi_i - L_a\sin(\lambda_i+\theta_a)} \quad (31)$$

After achieving the transmission matrix $T$, we combine Eqs.(14)(18) for obtaining the torques of phalange joints as follows,

$$\tau'_1 = X_1 \tau_a$$
$$\tau'_2 = X_2 \tau_a - K_2 \Delta\theta_2$$
$$\tau'_3 = X_3 \tau_a - K_3 \Delta\theta_3 \quad (32)$$

Therefore, the relationship of contact forces and torques from the base joints of the phalanges based on Eq.(18) can be expended as

$$\begin{bmatrix} k_1 & 0 & 0 \\ r_{12} & k_2 & 0 \\ r_{13} & r_{23} & k_3 \end{bmatrix}^T \begin{bmatrix} f_1 \\ f_2 \\ f_3 \end{bmatrix} = \begin{bmatrix} \tau'_1 \\ \tau'_2 \\ \tau'_3 \end{bmatrix} \quad (33)$$

The contact forces are yielded as

$$\begin{bmatrix} f_1 \\ f_2 \\ f_3 \end{bmatrix} = \begin{bmatrix} \frac{1}{k_1} & -\frac{r_{12}}{k_1 k_2} & \frac{r_{12}r_{23}-k_2 r_{13}}{k_1 k_2 k_3} \\ 0 & \frac{1}{k_2} & -\frac{r_{23}}{k_2 k_3} \\ 0 & 0 & \frac{1}{k_3} \end{bmatrix} \begin{bmatrix} \tau'_1 \\ \tau'_2 \\ \tau'_3 \end{bmatrix} \quad (34)$$

Furthermore, the following equation provides the relationship among the vector $\tau'$ of torques generated by the contact forces on the phalanges when all of them are in contact with the grasped object,

$$\tau' = \begin{bmatrix} \tau''_1 \\ \tau''_2 \\ \tau''_3 \end{bmatrix} = \begin{bmatrix} f_1 k_1 \\ f_2 k_2 \\ f_3 k_3 \end{bmatrix} = \begin{bmatrix} 1 & -\frac{r_{12}}{k_2} & \frac{r_{12}r_{23}-k_2 r_{13}}{k_2 k_3} \\ 0 & 1 & -\frac{r_{23}}{k_3} \\ 0 & 0 & 1 \end{bmatrix} \begin{bmatrix} \tau'_1 \\ \tau'_2 \\ \tau'_3 \end{bmatrix} \quad (35)$$

Thus,

$$\tau''_1 = \tau'_1 - \frac{r_{12}}{k_2}\tau'_2 + \frac{r_{12}r_{23}-k_2 r_{13}}{k_2 k_3}\tau'_3$$
$$\tau''_2 = \tau'_2 - \frac{r_{23}}{k_3}\tau'_3$$
$$\tau''_3 = \tau'_3$$

The above equation is valid if and only if $k_1 k_2 k_3 \neq 0$. If $k_1 k_2 k_3 = 0$, $J$ is a singular matrix. When the actuation torque brings the driving linkage to rotate $(\dot{\theta}_a \neq 0)$, $T$ must not be singular. It is well known that fewer than-full phalange grasps ($k_1 k_2 k_3 = 0$) can be stable [22]. However, a finger sometimes can realize fewer than full phalange grasps, which causes the singularity of $J$ so that Eq.(18) is not available. To address this issue ($k_1 k_2 k_3 = 0$) where the finger has fewer contact forces than the number of phalanges, we propose the following method to determine the distributions of contact forces. When the $i$-th phalanx does not contact the object, $f_i$ becomes zero so that the parameters $r_{ij}(j = 1,2,3)$ in Eq.(33) are not relevant. For obtaining the balance of contact forces(except for $f_i$), we delete the $i$-th column and $i$-th row in the matrix $J$ since all $r_{ij}$ and $r_{ji}, j = 1,2,3$ do not exist to $f_i$, and also remove $f_i$ and $\tau'_i$ from the contact force vector and torque vector of the joints of the phalanges, respectively. After deleting, the matrix $J$ is not singular. As a result, Equation (33) can be applied to achieving contact forces.

However, due to the coupling relationship among $\theta_1, \theta_2$ and $\theta_3$, it is not possible to establish accurate mapping relationships during the gripper's movement. Therefore, by classifying the grasping conditions and analyzing the structure of the gripper, the motion of the three joints of the gripper is decoupled. According to the order of the phalange contact between the phalange and the object, the grasping mode switching situation can be divided into three categories:

(1) When the distal phalange is the first to contact the object, the gripper will always keep parallel grasping, thereby remaining unchanged for $\theta_3$ while $\theta_1$ is reciprocal to $\theta_2$ with the proximal phalange moving.

(2) The middle phalange is the first to contact the object: the gripper keeps parallel motion before contacting the object; when the middle phalange touches the object, the gripper switches to the enveloping grasping mode with the distal phalange moving and the proximal joint stopping ($\theta_1$ and $\theta_2$ remain unchanged). We can achieve the mapping relationship between $\theta_i$ and $\theta_3$.

(3) The proximal phalange is the first to contact the object: when the proximal phalange contacts the object, the gripper switches to the enveloping grasping mode. The distal spring starts moving with the coupled angles $\theta_2$ and $\theta_3$ changing, and $\theta_1$ remains unchanged. After conducting a large number of grasping experiments, it can be found that the situation where the proximal phalange makes contact first is relatively rare. Other cases rarely occur in real grasping scenarios.

III. CONSTRUCTION OF FORCE SENSING

Here we develop an approach based on a combined current mathematical statistics model and LSTM model to obtain contact forces on phalanges for arriving at the force-feedback control without any force sensors.

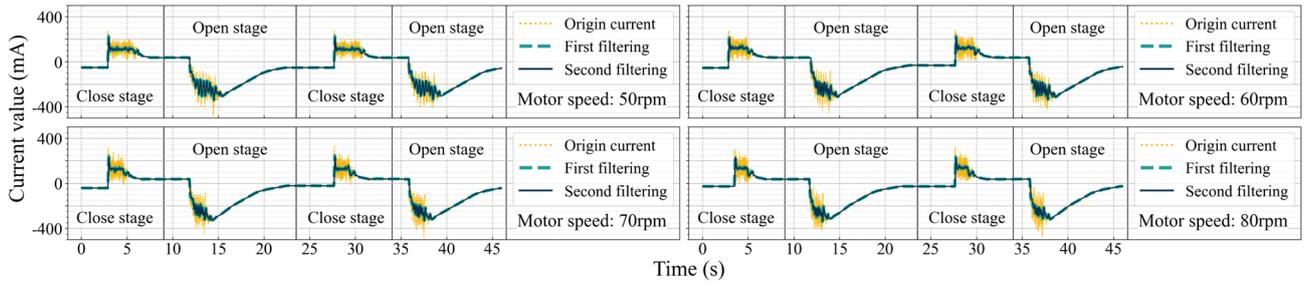

Figure. 5. The current changes of opening-closing actions for GL-Robot.

## A. Design of Semi-closed Loop Force Control

Semi-closed loop control strategy considers a comparison between the real output of the actuator system and expectation commands as a control-feedback. With an encoder sensor attached to the actuator, the current loop, velocity loop, and position loop are introduced to form a semi-closed-loop control system. Although it is not possible to obtain each joint angle due to the mutual under-actuation coupling, the movement parameters of joints in the finger can be decoupled through the classification of the grasping situation, as shown in Table I(all the tables in the appendix). Restating the previous formulations, we seek to find the joint position $\theta_i$ and velocity $\dot{\theta}_i$ from the position and velocity loops, respectively. Let $\tau_a$ denote the actuator torque achieved from the current loop, $\tau_a = AI$ ($A$ represents the torque constant of the motor and $I$ indicates the current). Computing the contact force of each phalange would seem to substitute $\theta_1$, $\theta_2$, $\theta_3$ achieved from $\theta_i$ into Eq.(34).

## B. Grasping Configuration via the Current Statistics

With the same in the actuator's speed, we can determine a grasping configuration of the gripper by the current changes owing to the loading varying. That is, the critical value of the joint angle $\theta_i$ in grasping mode switching can be predicted through the change of current. Thus, we pay close attention to the current fluctuation. The current of the actuator presents a Gaussian distribution curve concerning the expectation of the setting current. With the load increasing, the standard deviation $\sigma$ of the Gaussian distribution for the current curve will result in an improvement. As shown in Fig.5, it demonstrates the changes in the actuator current when the gripper opens and closes. To enable $\sigma$ to be stable, a two-stage filter is employed to converge the current. In particular, in the first stage, the median-value filter used significantly reduces the range of current fluctuations while sifting out the extreme values, as shown by the green dashed line in Fig.5; the second-stage filtering utilizes a mean value filter, which further treats the folded current curve as a more rounded one, as illustrated by the blue curve in Fig.5, allowing for the construction of feedback control system based on the filtered current. However, the mean-value filtering can result in an obvious delay. To relieve this issue, a delay array of units is used to record the position loop data to minimize the actuator position's reading error.

As a consequence, we can predict the critical transition status of the parallel grasp and enveloping grasp according to the remarkable change of the current, thereby obtaining the values of $\theta_1$, $\theta_2$ and $\theta_3$. In parallel grasping mode, the actual current change of the motor is small. The accurate mapping relationships between $\theta_i$ and $\theta_1$, $\theta_2$ can be built; when switching from parallel grasp to enveloping grasp mode, the current surges with the load increasing. It is considered as the gripper grasps the object. The mapping relationship between $\theta_i$ and $\theta_3$ can be accurately given. Figure 6 illustrates the relationship between the current changes and grasping configuration in the joint angle and grasping mode. The gripper maintains the parallel grasping mode without objects, repeating five-time grasping actions. In this case, the current framed by a rectangle dotted line fluctuates only in a small range(see Fig.6). For other 20-time grasps of objects whose dimensions gradually increase, it can be seen that there are two phases for the current fluctuations during the grasping procedures. In particular, the parallel motion phase without contacting the object has small current fluctuations. After the gripper contacts the object, the actuator current starts to surge and has significant fluctuations, which illustrates that the parallel grasping mode is switched to the enveloping case. Moreover, the amplitude of the surge becomes higher as the size of the grasped object increases.

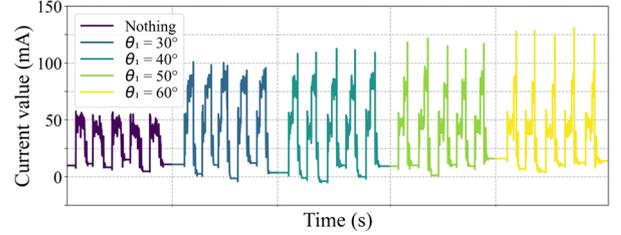

Figure. 6. The current changes for the gripper grasping objects with different dimensions.

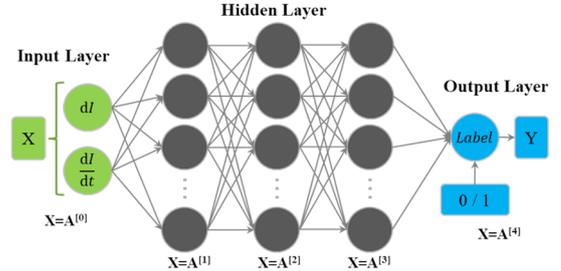

Figure. 7. The construction of the LSTM network model.

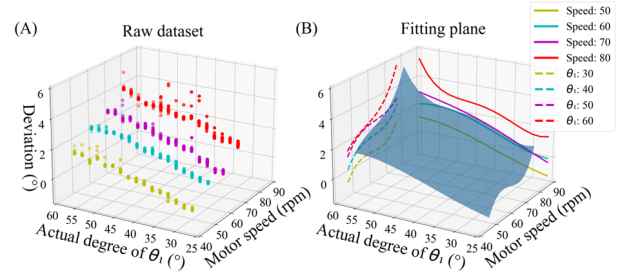

Figure. 8. The x-axis represents the speed of the motor, the y-axis represents the size of the grasping object, and the z-axis represents the deviation between $\theta_1$ predicted by LSTM and the actual $\theta_1$, and the data are fitted in the x-z plane and y-z plane sequentially, both of which can achieve better fitting results using quadratic polynomials and quartic polynomials, respectively, without overfitting, and the obtained fitted planes and the partial projections in the x-direction and y-direction(B).

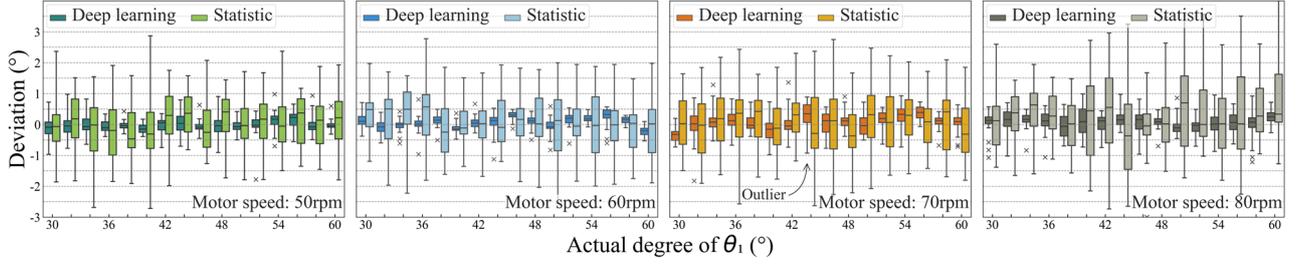

Figure. 9. The deviation results of $\theta_1$ predicted by the LSTM method for different objects at the different actuator's speeds.

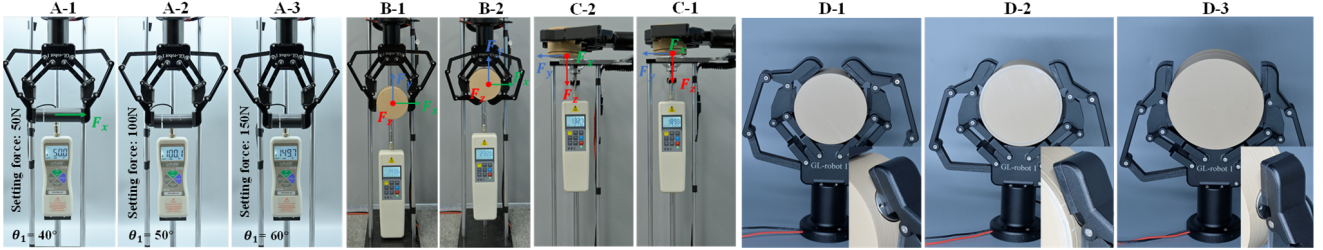

Figure. 10. Schematic diagrams of measuring the normal grasping force $F_x$ (A) with three different configurations in the first joint angle $\theta_1$ and setting force, the radial force $F_y$ (B) and the axial force $F_z$, respectively. Parallel grasps (B-1, C-1) and Enveloping grasps(B-2, C-2). Force sensing tests for multi-point enveloping grasps with different the first joint angles of GL-Robot (D).

### C. The Critical Angle Optimized by Deep Learning

We construct a deep-learning model to optimize the prediction $\theta_1$ (see Fig.7). The LSTM network is first adopted to predict $\theta_1$ during the grasping mode switching and then, the deviation between the LSTM-predicted $\theta_1$ and the actual $\theta_1$ is further compensated by a suitable mathematical model for improving the $\theta_1$ prediction accuracy.

LSTM is a special type of recurrent neural network model. By introducing memory cells, input gates, output gates, and forgetting gates, LSTM can more effectively cope with important events with longer intervals in time series; thus, it is suitable for LSTM to the random fluctuation of the current as time changes. The forgetting gate of the LSTM can combine the information of the previous and present current time steps, and comprehensively determine whether the intervals with large current fluctuations are forgotten or not. For training this model, the current information of the gripper to grasp 16 objects forming $\theta_1$ ranging from 30° to 60° is collected, and 100 sets of datasets are achieved for each object, totaling 1,600 sets of datasets. The collected information includes the actual value and the first-order derivative of the current as well as the actual angle of $\theta_1$. After data acquisition, the filtered current value and first-order derivative are inputted into the network. During the period of the model training, the time point is labeled as "0" when the actual angle of $\theta_1$ is smaller than the angle value formed by a grasping configuration of the gripper to grasp an object; otherwise, it is labeled as "1". The LSTM model takes the mark of the time point as the output, in particular, "0" indicating the parallel grasping mode and "1" meaning the enveloping grasping model.

The critical value $\theta_1$ directly predicted by the LSTM model still has a certain deviation from its actual value at the grasping mode switching due to the delay in the processing current using mean filtering, etc. Therefore, we employ a mathematical higher-degree polynomial fitting method to compensate for the deviation between the $\theta_1$ predicted by the LSTM and the actual $\theta_1$. Specifically, the gripper grasps 16 objects forming $\theta_1$ in the range from 30° to 60° at actuator speeds of 50rmp, 60rmp, 70rmp, and 80rpm, respectively. 50 sets of the critical value $\theta_1$ predicted by the LSTM model are achieved at each speed for each object, totaling 3,200 sets of data. As shown in Fig.8, the critical values of $\theta_1$ predicted in scenarios where GL-Robot grasps the same object at different actuator's speeds and different objects at the same actuator's speed are fitted by the quadratic polynomials and quartic polynomials, respectively.

We first verify the prediction accuracy of $\theta_1$ for different objects at the same speed. GL-Robot grasps 16 different objects, making $\theta_1$ from 30° to 60° with a step length of 2° at the actuator speed of 60rpm while each object is grasped 30 times. As shown in Fig.9 and Table II, the average prediction deviation between the predicted value and the actual value is within ±0.7° and the maximum deviation does not exceed 1.3°. Further, we re-carry out the same experiments for other speeds to verify the LSTM-based method is better than the statistical method in predicting $\theta_1$. Secondly, experiments are conducted to evaluate the performance of the proposed LSTM-based method in the $\theta_1$ prediction accuracy at different speeds. GL-Robot grasps the same object that makes $\theta_1$ to be 60° at the motor speeds from 50rpm to 80rpm with a step interval of 5rpm as an example under each speed for 30 times. The average deviation between the predicted value and the actual value is within ±0.8°, and the maximum deviation is not more than 1.1°.

In summary, we employ the LSTM network and mathematical fitting compensating method to achieve an improvement in the prediction accuracy of $\theta_1$ at grasping mode switching, relieving the effects on the delay caused by the mean-filtering processing currents, mechanical structure and the length of the mode-switching transition intervals.

## IV. EXPERIMENTS AND DISCUSSIONS

A series of robot grasping experiments in real scenarios are conducted to evaluate the performance of GL-Robot.

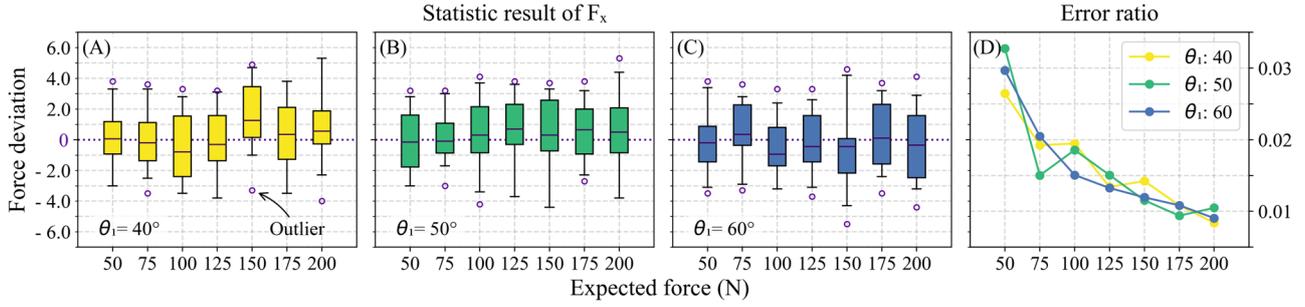

Figure. 11. Deviation distributions of testing the normal forces applied by the gripper with the first joint angle 40°, 50° and 60°(A,B,C) and the corresponding average deviation results (D).

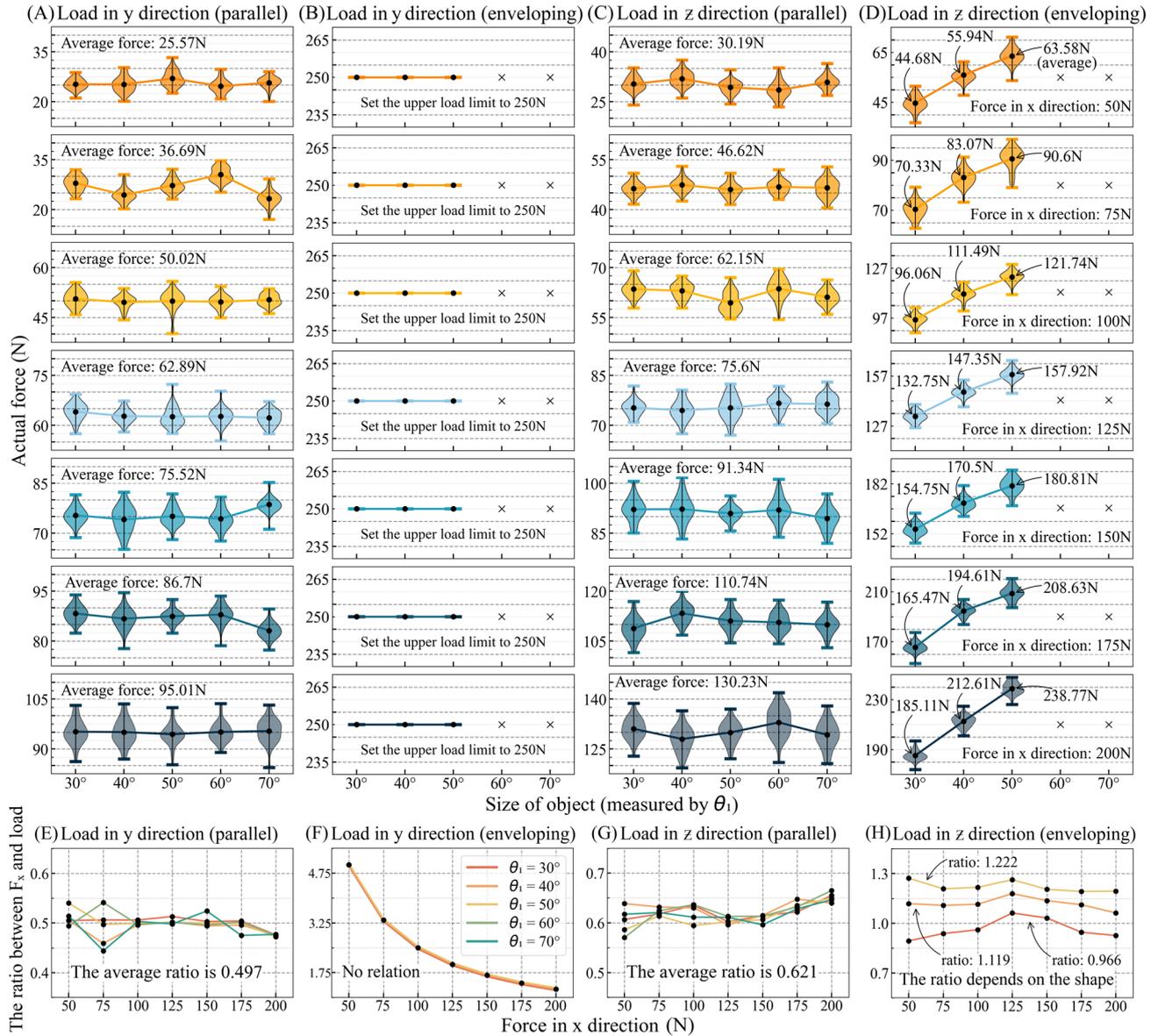

Figure. 12. Deviation distributions of testing the radial forces and axial forces applied by the gripper (A,B,C,D) and the ratio of the setting normal force and the radial load, axial load (E,F,G,H).

## A. Evaluation of Quantitative Indicators

In terms of the quantitative indicators (grasping payload/force sensing precision/stability/stiffness and large dimension ranges of objects), we primarily test the grasping payload/force sensing precision and the dimension ranges of objects to be grasped. As for the grasping stability indicator, our gripper can offer more stable grasps since it is easier to construct form-closure compared to a gripper with two 2-joint fingers. The grasping stiffness is tightly associated with the gripper material. Aluminum is used for building this gripper.

### 1) Grasping force

We conducted experiments on the grasping forces to verify the precision of force sensing by the proposed MS-LSTM model and the corresponding payload capacity of GL-Robot. The maximum grasping force can be in general deemed as the payload. According to the attitude of the gripper and the contact force direction, the grasping force is categorized into three directions, i.e., the normal force $F_x$, the radial force $F_y$ as well as the axial force $F_z$. An experimental setup is that GL-Robot is vertically placed on a designed platform consisting of a tension meter and a few 3D-printed objects, as shown in Fig.10. To achieve the maximum grasping forces (payloads), after the initial grip, we enable the actuator to be active all the time while the gripper holds an object.

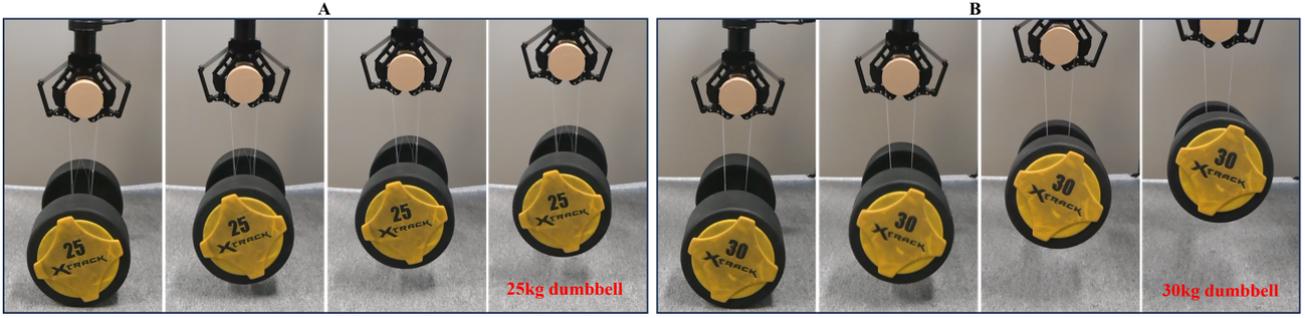

Figure. 13. Grasping dumbbells with 25kg (A) and 30kg (B) to test the radial force.

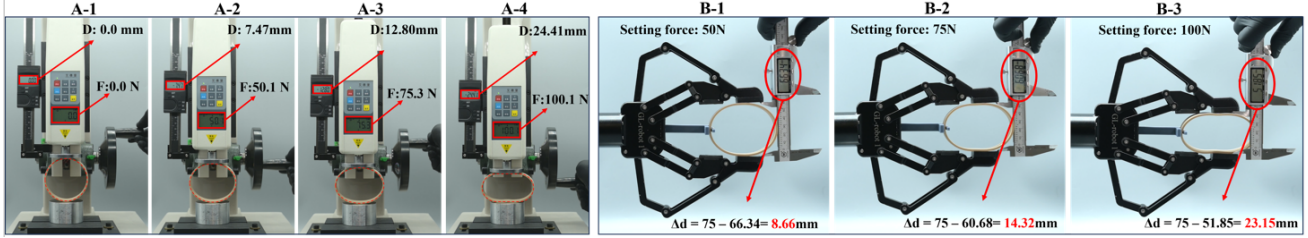

Figure. 14. Sampling the data of the force applied by dynameter to an elastic ring and the deformation of the ring (A). The grasping forces with 50N, 75N, 100N from GL-Robot and the corresponding deformations of the ring(B). $\Delta d$ represents the deformation of the ring.

*a) Determining the normal grasping force $F_x$*

In the experiment, we evaluate the performance of the proposed force-sensing method and the normal grasping payload. An off-the-shelf force sensor is applied to measure the normal grasping force $F_x$ that is dominant in the parallel grasps, where the center of the force sensor is always in contact with the midpoint of the distal phalange, i.e., at $k_3 = \frac{L_3}{2}$ (see Fig.10). Different grasping forces from 50N to 200N with a step size of 25N are set for grasping three objects that allow the first joint angle of the gripper $\theta_1$ to 40°, 50° and 60°, respectively. The statistical results of force sensing precision are shown in Fig.11 and Table III. The average deviation and the maximum deviation of the normal grasping forces are less than 1.71N and 6.5N, respectively. The average deviation rate is just 1.6%. As the setting force becomes larger, the precision of the grasping force sensing improves. The maximum normal grasping force of the gripper in some poses can be more than 340N, as shown in Fig.15 and Table V. Additionally, we have investigated force-sensing cases when GL-Robot holds an object with more than 2-point contact. A dynamometer installed on a cylinder tests the accuracy of force-sensing for the multi-point enveloping grasps (see Fig.10-D). We set different grasping forces and conducted 30 trials for objects with $\theta_1$ angles of 50°, 40°, and 30°, respectively. The results, presented in Table IV, indicate that the average error and error rate of force sensing for the multi-point enveloping grasps is less than 2.9N and 2.15%, respectively.

*b) Determining the radial force $F_y$ and the axial force $F_z$*

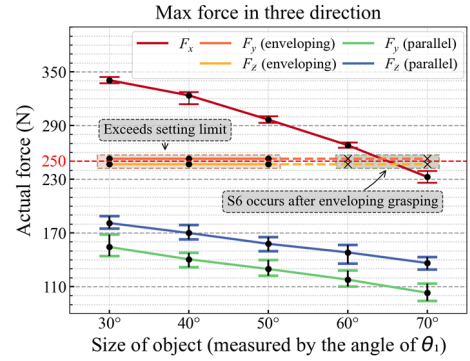

Figure. 15. Schematic diagram of the maximum grasping forces (payload) applied by the gripper with different angles of the first joint in the three directions.

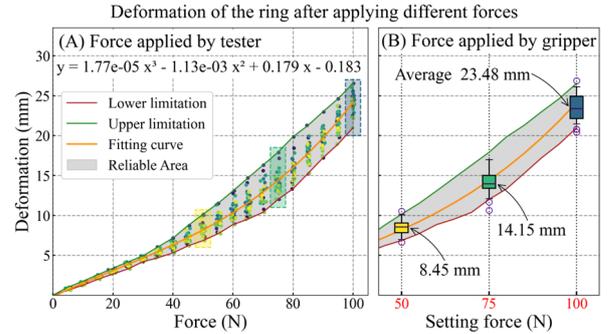

Figure. 16. Sampling the data of the relationship between the force applied to the ring and the deformation of the ring (A) and the relationship of the grasping force and the corresponding deformation being in agreement with the fitted function (B).

The object is firmly attached to the dynamometer so that an external force can be applied to pull the object out from the gripper. After the object is stably grasped by the corresponding force, the dynamometer is slowly moved downward through the rear screw slide to increase the tension in the y-direction and the z-direction (see Fig.10-B, C). When the object is at a standstill, the static friction between the grasped object and the gripper gradually increases as the dynamometer moves downward. Once the grasped object and the gripper produce relative displacement,

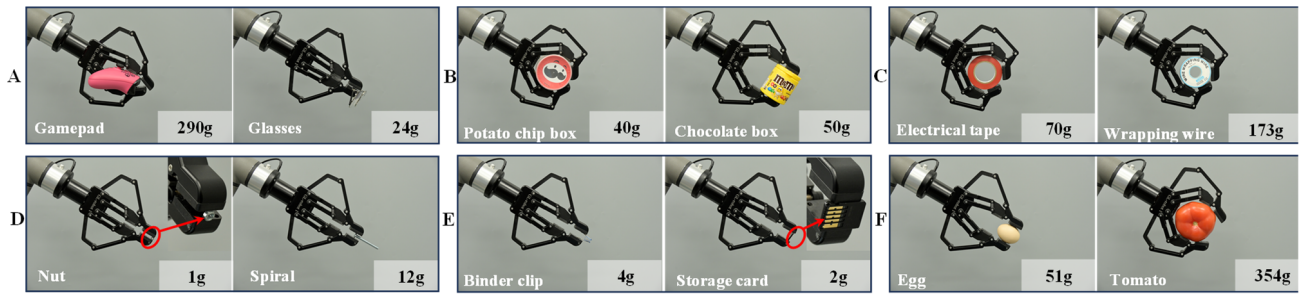

Figure. 17. Several demonstrations to verify the performances of the pinch-lifting gripper installed on a manipulator. Grasping objects with irregular shapes(A); Grasping objects with regular shapes(B); Grasping tools in the enveloping grasping mode(C); Grasping tools in the parallel grasping mode(D); Precision grasps(E); Grasping fragile objects (F). Note that supplemented cases are illustrated in the appendix.

the reader of the dynamometer keeps the same, which is considered as the payload.

In terms of statistical results illustrated in Fig.12 and Table VI/VII, it can be seen that both radial and axial forces are positively correlated with the set normal forces in the parallel grasping mode, with average correlation coefficients of 0.497 and 0.621, respectively. Almost all the coefficients are located in a narrow scope. In the enveloping grasping mode, although the axial load is related to the attitude formed by the gripper joints, the average correlation coefficients still keep in a small range from about 0.966 to 1.222 for different sizes of objects that allows the first joint angle of the gripper to 30°, 40° and 50°, respectively. Thus, it indirectly illustrates that the proposed force-sensing method has good performance.

As for the enveloping grasping mode, the axial load is not only related to the size of the gripping force, but also in connection with the attitude formed by the joints of the gripper, while in the radial gripping, as long as the gripper forms a type of closure with the grasped object, the load is only determined with the gripper's structural strength, which is not related to the magnitude of the grasping force. Two 3-phalange fingers and the palm take substantial contact force, undergoing wide and even pressure distribution, which indicates that squeezing out the object is unusual for an enveloping grasp of GL-Robot. Indeed, during the experimental periods, we recorded the peak radial force surpassing approximately 350N while the gripper stably held the object. However, the force magnitude should be measured once the object breaks away from the gripper. To prevent damage to the gripper's structure, the maximum grasping force is set at 250N, as shown in Fig.15. Thus, 250N is regarded as the maximum radial force of GL-Robot. Moreover, when the gripper performs the parallel grasps, its radial and axial payloads are up to 150N (approx. 15kg) and 180N (approx. 18kg), respectively.

*2) Grasping with the force-feedback control*

We conduct experiments by comparing deformations of an elastic ring after being forced from a dynamometer, intuitively verifying the effectiveness of the proposed force-sensing method in practical force-feedback control applications (see Fig.14). The relationship between the force applied to the elastic and the deformation of the elastic ring needs to be built. The forces from 0 to 100N with the step of 5N are applied to the ring, and the corresponding deformations are recorded. Then, we use a mathematical function to fit the relationship between the force and deformation, as shown in Table VIII and Fig.16(A). After completing the above standard sampling, the gripper grasps the ring with the set force 30 times at 50N, 75N, and 100N, and the width of the deformed ring is measured (see Fig.14). It can be found that the deformation degree of the ring under the corresponding grasping force is within the standard range obtained by the dynamometer test, which further proves the effectiveness of the gripper force control by the proposed force-sensing method.

*3) Dimension ranges of objects to be grasped*

To achieve the dimension ranges of objects to be grasped, we carry out tests of GL-Robot to grasp objects with various dimensions. GL-Robot can grasp a coin with a thickness of 1.85mm in the parallel grasping mode and a cubic object with a length of 125mm in the enveloping grasping mode. Therefore, GL-Robot can handle grasps of shell objects and some objects whose dimensions are almost equal to the maximum opening size of the gripper.

### B. Evaluation of the Capability of Grasping Objects

A series of experimental tests are conducted to demonstrate the versatility and robustness of GL-Robot on a variety of objects (see Fig.17).

To evaluate the versatility of GL-Robot, we enable the gripper to grasp objects of varying size, shape, texture, weight, and softness. Figure 17 shows the cases of grasping various irregularly shaped objects. When grasping an object with corners such as glasses, the contact may occupy only a small portion. In this case, GL-Robot is still able to present superior performance. Figure 17 indicates that the gripper uses the parallel grasping mode to pick up small objects while relatively big objects can be grasped via the enveloping grasps. GL-Robot can handle tools, which are potentially used in robotic dexterous manipulation. Owing to the under-actuated property, GL-Robot conforms with the irregular shapes of tools and contributes toward grasping stability. However, when attempting to grasp a paper on a table, GL-Robot fails, posing disadvantages for shell types of items from a platform.

### C. Discussions

Here we have discussions regarding the performance of the proposed GL-Robot.

As depicted in Fig.8, the surface fitting method incorporates all data collected in practical environments, encompassing screw gap errors, shaft-pin fit errors, encoder specifications, and other systematic errors as a unified entity. For instance, the encoder specifications, including resolution and accuracy, are integrally accounted for during the surface fitting process. The input data from real-world grasping scenarios are treated as outcomes influenced by the aforementioned factors. Consequently, there is no need to separately assess their impacts.

We discuss the potential failures regarding force sensing. It is a quite difficult challenge for sensing contact forces without any force sensors in complex grasping scenarios where GL-Robot grasps some irregular-shaped objects (see Fig.17-A:Gamepad). This is a genuine problem that may be solved by new machine learning-based approaches and specially designed under-actuated gripper. It is a promising approach to force estimation for the proposed GL-Robot by learning-from-simulation(LFS). LFS refers to a method of learning where instead of directly interacting with the real world, an agent learns from simulated environments or data generated by simulations. As for the LFS, GL-Robot typically grasps objects, interacting with a virtual environment or simulator that mimics aspects of the real world. Simulations can be used to generate vast amounts of training data, which can be valuable for training the proposed LSTM model.

GL-Robot exhibits a maximum grasping force (payload) of over 105N in the normal direction, surpassing that of Robotiq85. Furthermore, the maximum grasping force of GL-Robot is respectively seven and five times larger than that of Robotiq85 in the radial and axial directions, as shown in Table IX. The state-of-the-art under-actuated grippers examined in this paper lack the inherent ability to sense force without additional force sensors. Robotiq85 relies solely on adjusting the actuator's power to regulate the grasping force. However, according to published data, it has approximately 10% estimation errors, exceeding the less than 3% error rate demonstrated by GL-Robot. Consequently, in the context of force-feedback control, GL-Robot presents superior performance compared to Robotiq85. Furthermore, GL-Robot, equipped with two 3-joint fingers, offers enhanced grasping compliance and stability. This is attributed to its increased ease in achieving form-closure compared to Robotiq85 with two 2-joint fingers. GL-Robot, characterized by an increased number of phalanges and a broadened opening width, possesses a superior capability to grasp objects across a broader range of dimensions compared to the Robotiq85. Overall, given its exceptional performance and its high cost-effectiveness, we deem GL-Robot to be a viable candidate for potential commercialization.

## V. Conclusion

In summary, this article mainly focuses on the construction of a novel 2-finger 6-Dof under-actuated gripper with force sensing via deep learning. The experimental results illustrate that GL-Robot has impressive performances in grasping quantitative indicators, significantly outperforming Robotiq85 which is deemed as a baseline, which is potentially commercialized. As future directions, it is very interesting to investigate the geometry parameter optimization of the gripper and the gripper's material.

ACKNOWLEDGEMENT

This research was supported by the National Natural Science Foundation of China (the Youth Program: 509109-N72401), Robotics Institute of Zhejiang University(109107-I2170Q/033) and the National High-level Youth Talent Project (588020-X42306/008).